\documentclass[10pt,twocolumn,letterpaper]{article}

\usepackage{cvpr}              

\usepackage[accsupp]{axessibility} 

\usepackage{graphicx}
\usepackage{amsmath}
\usepackage{amssymb}
\usepackage{booktabs}
\usepackage{multirow}
\usepackage[pagebackref,breaklinks,colorlinks]{hyperref}

\newcommand{\jk}[1]{#1}
\newcommand{\Hu}[1]{#1}

\newcommand{\heading}[1]{\noindent\textbf{#1}}
\usepackage{pifont}
\newcommand{\cmark}{\ding{51}}%
\usepackage[capitalize]{cleveref}
\crefname{section}{Sec.}{Secs.}
\Crefname{section}{Section}{Sections}
\Crefname{table}{Table}{Tables}
\crefname{table}{Tab.}{Tabs.}

\makeatletter
\def\@fnsymbol#1{\ensuremath{\ifcase#1\or \dagger\or \ddagger\or
\mathsection\or \mathparagraph\or \|\or **\or \dagger\dagger
\or \ddagger\ddagger \else\@ctrerr\fi}}

\newcommand{\printfnsymbol}[1]{%
  \textsuperscript{\@fnsymbol{#1}}%
}
\makeatother

\begin{document}

\title{Spatial-Temporal Space Hand-in-Hand:\\
Spatial-Temporal Video Super-Resolution via Cycle-Projected Mutual Learning}

\author{Mengshun Hu${^{1,2}}$\thanks{ Equal Contribution} \quad Kui Jiang${^{1,2}}$\printfnsymbol{1} \quad Liang Liao${^{3}}$ \quad Jing Xiao${^{1,2}}$ \quad Junjun Jiang${^{4}}$ \quad Zheng Wang${^{1,2}}$\thanks{ Corresponding Author} \\
\normalsize{${^1}$National Engineering Research Center for Multimedia Software, Institute of Artificial Intelligence, School of} \\ 
\normalsize{Computer Science, Wuhan University ${^2}$Hubei Key Laboratory of Multimedia and Network Communication Engineering}\\
\normalsize{${^3}$Nanyang Technological University \quad ${^4}$Harbin Institute of Technology }
}

\maketitle

\begin{abstract}
Spatial-Temporal Video Super-Resolution (ST-VSR) aims to generate \jk{super-resolved} 
videos with higher resolution (HR) and higher frame rate (HFR). Quite intuitively, pioneering two-stage based methods \jk{complete ST-VSR by} 
directly \jk{combining} two sub-tasks: Spatial Video Super-Resolution (S-VSR) and Temporal Video Super-Resolution (T-VSR) \jk{but ignore the reciprocal relations among them. Specifically, }
1) T-VSR to S-VSR: temporal correlations \jk{help accurate spatial detail representation with more clues;} 
2) S-VSR to T-VSR: abundant spatial information contributes to the refinement of temporal prediction. To this end, we propose a one-stage based Cycle-projected Mutual learning network (CycMu-Net) for ST-VSR, which makes full use of spatial-temporal correlations via the mutual learning between S-VSR and T-VSR. Specifically, 
we propose to exploit the mutual information among them via iterative up-and-down projections, where the spatial and temporal features are fully fused and distilled, helping the high-quality video reconstruction. 
\jk{Besides extensive experiments on benchmark datasets, we also compare our proposed CycMu-Net with S-VSR and T-VSR tasks, demonstrating that our method significantly outperforms state-of-the-art methods. Codes are publicly available at}: 
\url{https://github.com/hhhhhumengshun/CycMuNet}.
\end{abstract}

\section{Introduction}
\label{sec:intro}


\tabcolsep=0.5pt
\begin{figure*}[htb]
     \centering
     \includegraphics[width=0.97\textwidth]{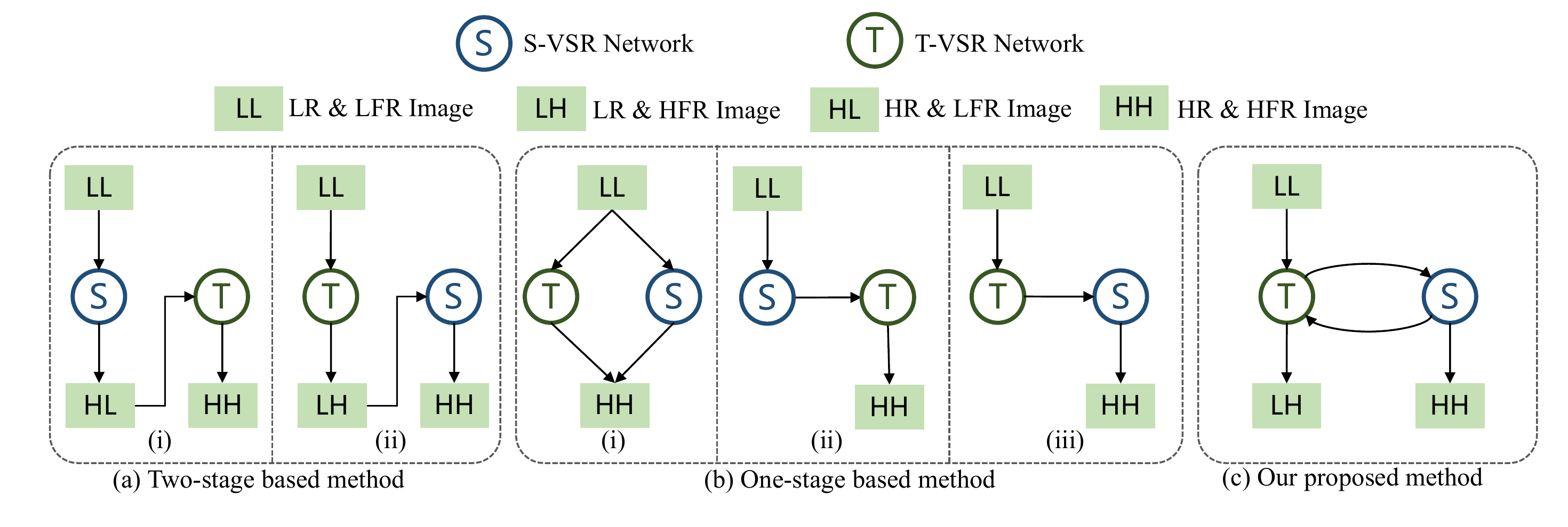}
    \vspace{-5mm}
     \caption{Different schemes for ST-VSR. (a) Two-stage based methods: (i) they perform ST-VSR task by independently using the advanced S-VSR methods and then T-VSR methods or vice versa (ii). (b) One-stage based method: they unify S-VSR and T-VSR tasks into \jk{one model with parallel or cascaded manners without considering the mutual relations between S-VSR and T-VSR.} 
     (c) \jk{Mutual} 
     method: \textbf{Our method makes full use of the mutual relations via mutual learning between S-VSR and T-VSR.}}
     \label{motivation}
     \vspace{-5mm}
\end{figure*}

Spatial-temporal video super-resolution (ST-VSR) aims to produce the high-resolution (HR) and high-frame-rate (HFR) 
video sequences from the given low-resolution (LR) and low-frame-rate (LFR) input. This task has \jk{drawn} 
great attention 
due to its popular applications~\cite{xiang2020zooming,kim2020fisr,kang2020deep}, \jk{including} HR slow-motion generation, movie production, high-definition television upgrades, \emph{etc}. 
\jk{Great success has been recently achieved in ST-VSR tasks, as illustrated in Figure~\ref{motivation}(a), which can be roughly divided into two categories: two-stage and one-stage based methods.} 
\jk{The former decomposes it into two sequential sub-tasks: spatial video super-resolution (S-VSR) and temporal video super-resolution (T-VSR), which are individually completed with image/video super-resolution technologies~\cite{haris2019deep,zhang2018image,wang2019edvr} and video frame interpolation technologies~\cite{jiang2018super,niklaus2017videob}.} 
\jk{However, more spatial information generated by the S-VSR task can be used for the refinement of temporal prediction, while more temporal information predicted by the T-VSR task can be used to facilitate the reconstruction of spatial details. As a result, the two-stage based approaches are far from producing satisfied predictions due to lacking the ability to mutually explore the coupled correlations between S-VSR and T-VSR.}

Recently, integrating these two sub-tasks into a unified framework with a one-stage process becomes more popular. Naturally, based on the \jk{parallel or serially} processing modes (Figure~\ref{motivation}(b) (i) for parallel process and (ii)(iii) for serial process), diverse and effective schemes have been developed~\cite{kim2020fisr,kang2020deep,chan2020understanding,chan2021basicvsr,xiang2020zooming,xu2021temporal}.
Unfortunately, the parallel  methods~\cite{kim2020fisr,kang2020deep} barely consider the coupled correlations between the two sub-tasks, while the serial methods~\cite{xiang2020zooming,xu2021temporal} fail to fully exploit mutual relations since they only focus on the unilateral relationship, such as ``T-to-S" or ``S-to-T". In particular, the unilateral learning will accumulate reconstruction errors, which we define as cross-space (spatial and temporal spaces) errors, consequently leading to obvious aliasing effect in \jk{super-resolved} 
results.

For thorough utilization of spatial and temporal information, \jk{we propose to promote the one-stage method with mutual learning}, and 
devise a novel 
cycle-projected mutual learning network (CycMu-Net) for ST-VSR. As shown in Figure \ref{motivation}(c), the philosophy of CycMu-Net is to explore the mutual relations and achieve the spatial-temporal fusion to eliminate the cross-space errors. Specifically, the key part of CycMu-Net is the iterative up-and-down projection units between the spatial and temporal embedding spaces, involving a process of aggregating temporal relations to achieve an accurate representation of spatial details, and a feedback refinement of temporal information via the updated spatial prediction. 
We validate the proposed CycMu-Net on the ST-VSR task and its two sub-tasks, 
\jk{involving} S-VSR and T-VSR. Experimental results demonstrate that CycMu-Net achieves appealing improvements over the SOTA methods on all tasks. 
Our contributions are summarized as follows:
1) We propose a novel one-stage based cycle-projected mutual learning network (CycMu-Net) for spatial-temporal video super-resolution, which can make full use of the coupled spatial-temporal correlations via mutual learning between S-VSR and T-VSR.

2) \jk{To eliminate the cross-space errors and promote the inference accuracy, }
we devise iterative up-and-down projection units to exploit the mutual information between S-VSR and T-VSR \jk{for a better spatial-temporal fusion}. In these units, more spatial information are provided for the refinement of temporal prediction while temporal correlations are used to promote texture and detail reconstruction.

3) \jk{We conduct extensive 
experiments on ST-VSR, S-VSR and T-VSR tasks for a comprehensive evaluation, 
showing that our} method performs well against the state-of-the-art methods. 

\section{Related Work}
\label{sec:rea}

\subsection{Spatial Video Super-Resolution}
S-VSR aims to super-resolve LR frames to HR frames \jk{with temporal alignment and spatial fusion.} 
Thus, the key to this task lies in 
\jk{fully exploiting} temporal correlations among multiple frames. Some methods perform temporal alignment using explicit motion estimation (\emph{e.g.}, optical flow) and then fuse all aligned reference frames for S-VSR~\cite{caballero2017real,tao2017detail,sajjadi2018frame,wang2018learning,xue2019video,bao2019memc}. However, optical flow estimation is error-prone, \jk{which may degrade} the S-VSR performance~\cite{li2020mucan}. To address this issue, some methods propose \jk{to apply deformable convolution to sample more spatial pixels based on multiple motion offsets~\cite{dai2017deformable,zhu2019deformable} for implicit alignment}~\cite{wang2019edvr,tian2020tdan,chan2020understanding}. \jk{It is effective but time-consuming, since the alignment is required for all reference frames each time when super-resolving the target frame.} 
\jk{Other researchers propose to explore the global temporal correlations with recurrent networks that propagate inter-frame information forward and backward independently}~\cite{xiang2020zooming,xu2021temporal,huang2017video,chan2021basicvsr}. However, extra motion estimation networks are still required to assist the recurrent network based S-VSR approach in dealing with large and complex motions~\cite{xiang2020zooming,xu2021temporal}. 


\subsection{Temporal Video Super-Resolution}
T-VSR (\emph{i.e.}, video frame interpolation) aims to generate the non-existent intermediate frame between two consecutive frames. The key to this task is to find correspondences between consecutive frames 
to synthesize intermediate frames. The popular T-VSR methods mainly fall into two categories: kernel-based and flow-based methods. The former implicitly \jk{aligns} the input frames by \jk{learning} 
the dynamic convolution kernels, which are used to resample the input frames to produce intermediate frames~\cite{cheng2021multiple,niklaus2017videoa,gui2020featureflow,niklaus2017videob,lee2020adacof,shi2021video}. 
\jk{Due to only 
resampling the local neighborhood patches, the aforementioned methods usually lead to ambiguous results.} \jk{By contrast, the latter} 
first estimates bidirectional optical flows between two consecutive frames \jk{and then warps to synthesize the intermediate frames 
based on the predicted optical flows}~\cite{bao2019depth,bao2019memc,niklaus2020softmax,jiang2018super,niklaus2018context,hu2020motion,hu2021capturing}. 
\jk{While achieving} impressive progress, they rely heavily on the accuracy of current advanced optical flow algorithms~\cite{sun2018pwc,teed2020raft,hui2018liteflownet,ranjan2017optical}. 


\begin{figure*}[t]
     \centering
     \includegraphics[width=0.9\textwidth]{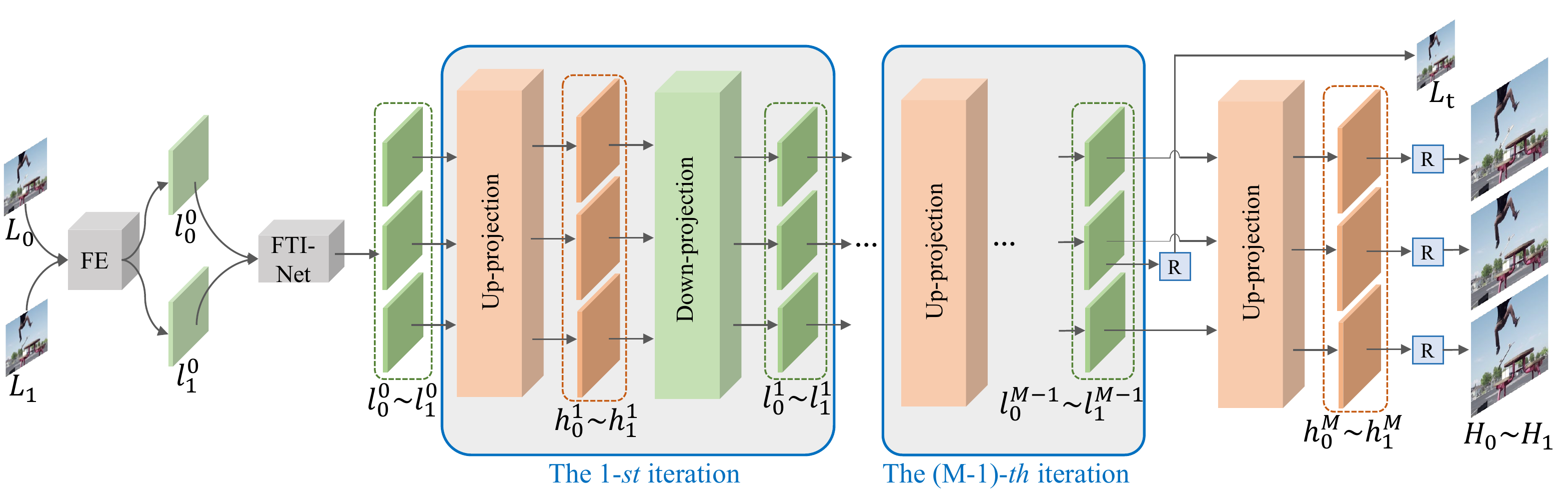}
          \vspace{-3mm}
     \caption{\textbf{Architecture of the proposed Cycle-projected Mutual learning network (CycMu-Net).} Given two LR input frames, we first extract representations from input frames by feature extractor (FE) and obtain an initialized intermediate representation by feature temporal interpolation network (FTI-Net). We then adopt mutual learning to exploit the mutual information between S-VSR and T-VSR and obtain $M$ 2 $\times$ HR and LR representations via $M$ up-projection units and $M-1$ down-projection units. Finally, we concatenate and feed the multiple 2$\times$ HR representations and LR representations into reconstruction network (R) to reconstruct corresponding HR images and LR intermediate frame, respectively.}
     \vspace{-5mm}
     \label{framework}
\end{figure*}

\subsection{Spatial-Temporal Video Super-Resolution}
\jk{ST-VSR technologies tend to} 
increase spatial and temporal resolution of LR and LFR videos~\cite{haris2020space,xiang2020zooming,xu2021temporal,kim2020fisr}. \jk{For example,} Shechtman \textit{et al.} \jk{adopt} a directional spatial-temporal smoothness regularization to constrain high spatial-temporal resolution video reconstruction~\cite{shechtman2002increasing}. Mudenagudi \textit{et al.} \cite{mudenagudi2010space} formulate their ST-VSR method as a posteriori-Markov Random Field~\cite{geman1984stochastic} and optimize it by achieving the Maximum of graph-cuts~\cite{boykov2001fast}. However, the above methods \jk{cost great computational consumption} 
and 
\jk{fail} to model complex spatial-temporal correlations. Recently, learning-based methods attempt to unify S-VSR and T-VSR into a single-stage framework for ST-VSR. Kim \textit{et al.} utilize a multi-scale U-net to learn ST-VSR based on a multi-scale spatial-temporal loss~\cite{kim2020fisr}. Haris \textit{et al.} propose to explore spatial-temporal correlations by a pre-trained optical flow model for frame interpolation and refinement~\cite{haris2020space}. Xiang \textit{et al} devise a unified framework to interpolate intermediate features by deformable convolution~\cite{wang2019edvr}, explored global temporal correlations by bidirectional deformable ConvLSTM~\cite{xingjian2015convolutional}, and finally reconstructed high spatial-temporal videos by a reconstruction network~\cite{xiang2020zooming}. Inspired by~\cite{xiang2020zooming}, Xu \textit{et al.} introduce a locally temporal feature comparison module to extract local motion cues in videos, achieving better performance on various datasets~\cite{xu2021temporal}. However, as shown in Figure \ref{motivation}(b), \jk{the mutual relations between S-VSR and T-VSR are under-explored, while leading to the accumulated reconstruction errors.}
To address this issue, we propose a cycle-projected mutual learning network \jk{that learns the spatial-temporal correlations via the iterative operation of spatial and temporal fusion (S-VSR and T-VSR) during the forward propagation and backward optimization.}

\subsection{\Hu{Mutual Learning}}

\Hu{Mutual learning is to make a pool of untrained students to learn collaboratively and teach each other for solving the task~\cite{zhang2018deep}. Dual-NMT utilizes mutual learning to teach two cross-lingual translation models  each other interactively machine translation~\cite{he2016dual}. Tanmay Batra \textit{et al.} propose to learn multiple models jointly and communicate object attributes each other for recognising the same set of object categories~\cite{batra2017cooperative}. Dong \textit{et al.} adopted this tool to exploit non-adjacent features for image dehazing by fusing features from different levels~\cite{dong2020multi}. The closest thing to our work is DBPN~\cite{haris2019deep}, which proposes utilize mutually iterative up- and down-sampling layers to learn nonlinear relationships between LR and HR images to guide the image SR task. Previous studies have validated the effectiveness of mutual learning techniques for low-level tasks~\cite{zhao2017iterative,dong2009nonlocal,dai2007bilateral,haris2020deep}. However, the existing methods tend to exploit the mutual learning to refine the mapping relations of different  scale spaces (``LR-to-HR" and ``HR-to-LR"). Inspired by them, we introduce a novel cycle-projected mutual learning mechanism to cooperatively characterise the spatial and temporal feature representations.}



\section{Cycle-Projected Mutual Learning  Network}
In this section, we first provide an overview of the proposed Cycle-projected Mutual learning network (CycMu-Net) for ST-VSR. \Hu{As shown in Figure~\ref{framework}}, given two LR input frames $L_{0}$ and $L_{1}$, our goal is to synthesize HR intermediate frame $H_{t}$ and the corresponding HR input frames $H_{0}$ and $H_{1}$ (2$\times$, 4$\times$, or 8$\times$). In addition, we also generate a LR frame $L_{t}$ as a intermediate result. The proposed CycMu-Net first extracts the representation from the input frames by a feature extractor (FE). To synthesize the initialized LR intermediate representation, we introduce a cascading multi-scale architecture as our feature temporal interpolation network (FTI-Net), 
\Hu{designed to learn bi-directional motion offsets to handle complex motions and interpolate intermediate representation by deformable convolution.} To make full use of the mutual relations (``T-to-S" or ``S-to-T")  between S-VSR and T-VSR, we adopt mutual learning that temporal correlations contribute to accurate spatial representations and updated spatial predictions refine temporal information 
\Hu{via} feedback, to eliminate the cross-space errors, which can be achieved via iterative up-projection units (UPUs) and down-projection units (DPUs). After several iterations, we obtain  multiple 
HR and LR representations and then concatenate them into the reconstruction network (R) to generate the corresponding HR images $H_{0}$, $H_{t}$ and $H_{1}$ (2$\times$, 4$\times$, or 8$\times$) and LR image $L_{t}$. 


\tabcolsep=0.5pt
\begin{figure}[t]
     \centering
     \includegraphics[width=1.0\linewidth]{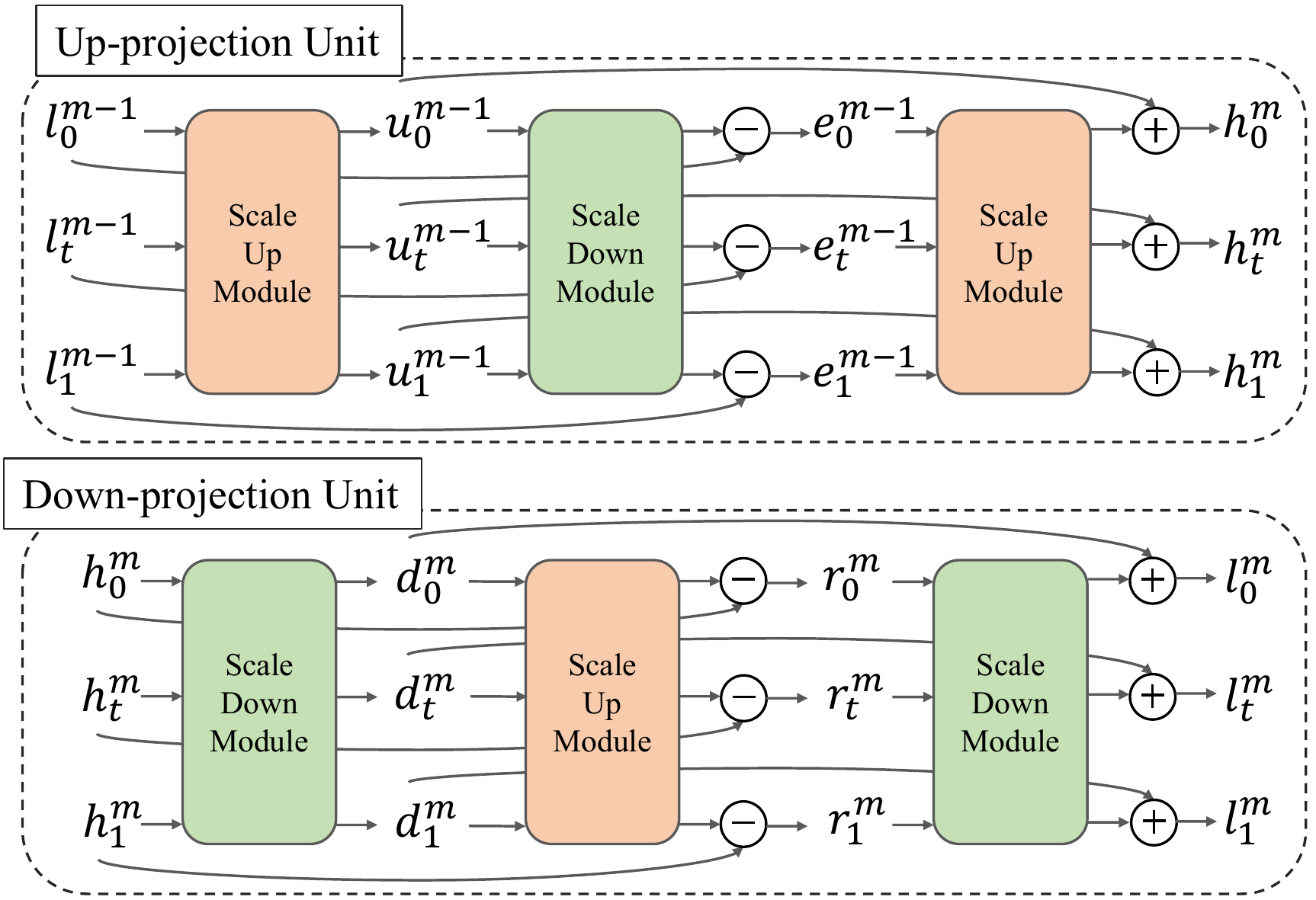}
     \vspace{-5mm}
     \caption{Illustration of the proposed \textbf{up-projection unit (UPU)} and \textbf{down-projection unit (DPU)} in the CycMu-Net.}          \vspace{-5mm}
     \label{framework1}
\end{figure}




\subsection{Cycle-Projected Mutual Learning}
Inspired by~\cite{haris2019deep} that adequately addressed the mutual dependencies of low- and high-resolution images via mutually connected up- and down-sampling layers, in this paper, we propose a new mutual learning model including iterative UPUs and DPUs to explore the mutual relations between S-VSR and T-VSR. In particular, temporal correlations provide more clues to compensate detailed spatial representation via UPUs while abundant spatial details are used to refine the temporal predictions via DPUs.

As shown in the top of Figure~\ref{framework1}, the UPU captures temporal correlations for S-VSR. We firstly project 
\Hu{previous} LR temporal representations $l_{0}^{m-1}$, $l_{t}^{m-1}$ and $l_{1}^{m-1}$ to corresponding HR representations $u_{0}^{m-1}$, $u_{t}^{m-1}$ and $u_{1}^{m-1}$ based on a scale up module, which can be described as follows:
\begin{align}
\begin{split}
 [u_{0}^{m-1},u_{t}^{m-1},u_{1}^{m-1}] = UP_0([l_{0}^{m-1},l_{t}^{m-1},l_{1}^{m-1}]),
\end{split}
\end{align}
where $UP_0(\cdot)$ denotes the scale up module. It first performs multi-frame progressive fusion by fusion resblocks~\cite{yi2019progressive}, which implicitly exploit intra-frame spatial correlations and inter-frame temporal correlations, then upsamples each feature by bilinear interpolation and 1$\times$1 convolution. $m=1,2...,M$ denotes the number of UPU.

Then we try to project the super-resolved representations back to LR representations and compute the corresponding residuals (errors) $e_{0}^{m-1}$, $e_{t}^{m-1}$ and $e_{1}^{m-1}$ between back-projected representations and original LR representations, respectively, which can be defined as follows:
\begin{align}
\begin{split}
[e_{0}^{m-1},e_{t}^{m-1},e_{1}^{m-1}] &= DN([u_{0}^{m-1},u_{t}^{m-1},u_{1}^{m-1}])\\ 
 &- [l_{0}^{m-1},l_{t}^{m-1},l_{1}^{m-1}],\\ 
\end{split}
\end{align}
where $DN(\cdot)$ denotes the scale down module. It first reduces the input to the original input resolution via 4$\times$4 convolution with stride 2, and then further implicitly explores intra-frame spatial correlations and inter-frame temporal correlations of LR representations by fusion resblocks~\cite{yi2019progressive}.

Finally, we project residual representations again back to HR representations (back-project) and eliminate the corresponding original super-resolved representations errors (cross-space errors) to obtain the final super-resolution outputs of the unit by
\begin{align}
\begin{split}
[h_{0}^{m},h_{t}^{m},h_{1}^{m}] &= UP_1([e_{0}^{m-1},e_{t}^{m-1},e_{1}^{m-1}])\\
 &+[u_{0}^{m-1},u_{t}^{m-1},u_{1}^{m-1}],  
\end{split}
\end{align}
where $UP_1(\cdot)$ denotes the scale up module.

As shown in the bottom of Figure~\ref{framework1}, the procedure for DPU is very similar, while its main role is to obtain refined LR temporal representations by projecting the previously updated HR representations, which can provide abundant spatial details. (Please refer to the supplementary materials for more details about formula proof, scale up module and scale down module)

\subsection{Spatial-Temporal Video Super-Resolution}
The overall framework of CycMu-Net is shown in Figure~\ref{framework}, consisting of the following sub-modules: feature extraction network, feature temporal interpolation network, multiple up-projection units, multiple down-projection units, and reconstruction network. Specifically, we extract representations among multiple frames via feature extraction network (FE) and interpolate the intermediate representations via the feature temporal interpolation network (FTI-Net). Then we use the proposed multiple UPUs and DPUs to obtain multiple LR and HR representations with the mutual learning. Finally, the reconstruction network (R) generates LR intermediate frame and HR intermediate frames by concatenating all LR and HR representations. Below we describe the details of each sub-module.

\heading{Feature temporal interpolation network.}
Deformable convolution~\cite{dai2017deformable,zhu2019deformable} has been shown to be effective for video frame interpolation~\cite{cheng2020video} and video super-resolution~\cite{tian2020tdan}. Some methods extended deformable convolution and explored a wider range of offsets by employing a multi-scale framework to handle feature alignment for small and large displacements~\cite{wang2019edvr,xiang2020zooming,xu2021temporal}. 
Inspired by them, we utilize a cascading multi-scale architecture for our feature temporal interpolation network (FTI-Net) to estimate the bi-directional motion offsets from input frames. Along with the motion offsets estimation, we adopt deformable convolution to interpolate forward and backward representations 
\Hu{from} the missing intermediate frames. To blend these two representations for obtaining an initial intermediate representation, we use the two learnable convolution kernels to estimate the weights, which can adaptively fuse the two representations according to their importance. (More details on FTI-Net are provided in the supplementary materials)

\heading{Reconstruction network.}
After the mutual relations between S-VSR and T-VSR are exploited 
\Hu{by the proposed iterative up-and-down projections,} we concatenate and feed multiple HR representations into convolution layers to reconstruct the corresponding HR frames. In addition, we also reconstruct a LR intermediate frame based on multiple LR representations. To optimize the whole CycMu-Net, we use a reconstruction loss function:
\begin{align}
\begin{split}
 {\mathcal{L}_{r}} &={\lambda}_{1} \rho({L}_{t}-L_{t}^{GT}) + {\lambda}_{2}\rho({H}_{t}-H_{t}^{GT}) \\
 &+ {\lambda}_{3}\rho({H}_{0}-H_{0}^{GT}) +  {\lambda}_{4}\rho({H}_{1}-H_{1}^{GT}),
\end{split}
\end{align}
where $L_{t}^{GT}$, $H_{0}^{GT}$, $H_{t}^{GT}$ and $H_{1}^{GT}$ refer to the corresponding ground-truth video frames. ${\rho}(x)= \sqrt{x^{2}+\omega ^{2}}$ is the Charbonnier penalty function~\cite{charbonnier1994two,lai2017deep}. We set the constant $\omega$ and weights $\lambda_{1}$, $\lambda_{2}$, $\lambda_{3}$ and $\lambda_{4}$ to $10^{-3}$, $1$, $1$, $0.5$ and $0.5$, respectively.

\subsection{Implementation Details}

We implement the proposed CycMu-Net using Pytorch 1.9 with four NVIDIA 2080Ti and optimize the model using AdaMax optimizer~\cite{kingma2014adam} with a momentum of 0.9. The batch size is set to 10 with image resolution of 64$\times$64. The initial learning rate is set to 4$\times${$10^{-4}$} and reduced by a factor of 10 every 20 epochs for a total of 70 epochs. We compare HR intermediate frame $H_{t}$ for the evaluation of ST-VSR. In addition, we also compare our proposed CycMu-Net with S-VSR and T-VSR methods, where \Hu{4$\times$} HR frame $H_{0}$ and LR intermediate frame $L_{t}$ are used for the evaluations of S-VSR and T-VSR, respectively.

\section{Experimental Results}

\subsection{Datasets and Metrics}

\heading{Vimeo90k~\cite{xue2019video}.}
We use Vimeo90K dataset to train our proposed CycMu-Net. This dataset consists of many triplets with different scenes from 14,777 video clips with image resolution of 448$\times$256. Among them, 51,312 triplets and 3,782 triplets are used for training and testing, respectively. In order to increase the diversity of data, we use horizontal and vertical flipping or reverse the order of input frames for data augmentation. For a fair comparison with other algorithms during training, we downscale to original images to 64$\times$64 with Bicubic interpolation for 2$\times$ and 4$\times$ SR, and downscaled to original images to 32$\times$32 with Bicubic interpolation for 8$\times$ SR.

\begin{table*}[htb]
\setlength\tabcolsep{4pt}
\centering
\resizebox{1.0\textwidth}{!}
{
\smallskip\begin{tabular}{cc|cccccc|cccccc|cccccc|c}
\hline
\multicolumn{1}{c}{T-VSR}&\multicolumn{1}{c}{S-VSR}& \multicolumn{3}{|c}{UCF101} & \multicolumn{3}{c}{Vimeo90K}& \multicolumn{3}{|c}{UCF101} & \multicolumn{3}{c}{Vimeo90K}& \multicolumn{3}{|c}{UCF101} & \multicolumn{3}{c|}{Vimeo90K}&{Parameters}\cr
Method&Method&PSNR&SSIM&IE&PSNR&SSIM&IE&PSNR&SSIM&IE&PSNR&SSIM&IE&PSNR&SSIM&IE&PSNR&SSIM&IE&(millions)\cr \hline
SepConv~\cite{niklaus2017videob}&Bicubic&29.988&0.944&4.531&30.628&0.937&4.234&26.189&0.874&7.154&27.287&0.866&6.582&22.877&0.779&11.201&24.181&0.782&9.989&21.7 \cr
SepConv~\cite{niklaus2017videob}&DBPN~\cite{haris2019deep}&32.041&0.958&3.729&32.179&0.955&3.415&28.380&0.915&5.573&28.969&0.903&5.268&25.135&0.845&8.298&26.016&0.834&7.717&21.7+10.4 \cr
SepConv~\cite{niklaus2017videob}&RBPN~\cite{haris2019recurrent}&31.859&0.957&3.795&32.377&0.958&3.300&28.650&0.920&5.400&29.507&0.914&4.912&25.323&0.823&8.067&26.409&0.846&7.275&21.7+12.7\cr
SepConv~\cite{niklaus2017videob}&EDVR~\cite{wang2019edvr}&---&---&---&---&---&---&28.650&0.920&5.388&29.481&0.914&4.909&---&---&---&---&---&---&21.7+20.7\cr
\hline
AdaCoF~\cite{lee2020adacof}&Bicubic&30.056&0.945&4.458&30.760&0.936&4.203&26.187&0.874&7.133&27.243&0.864&6.624&22.877&0.778&11.193&24.160&0.781&10.029&21.8\cr
AdaCoF~\cite{lee2020adacof}&DBPN~\cite{haris2019deep}&32.167&0.958&3.630&32.341&0.954&3.401&28.557&0.917&5.430&29.214&0.903&5.207&25.164&0.845&8.253&25.935&0.832&7.804&21.8+10.4\cr
AdaCoF~\cite{lee2020adacof}&RBPN~\cite{haris2019recurrent}&31.997&0.958&3.692&32.537&0.957&3.288&28.840&0.922&5.237&29.584&0.914&4.865&25.349&0.851&8.026&26.155&0.841&7.466&21.8+12.7\cr
AdaCoF~\cite{lee2020adacof}&EDVR~\cite{wang2019edvr}&---&---&---&---&---&---&28.848&0.923&5.226&29.700&0.916&4.810&---&---&---&---&---&---&21.8+20.7\cr
\hline
CAIN~\cite{choi2020channel}&Bicubic&29.931&0.941&4.627&30.578&0.931&4.412&25.987&0.865&7.456&26.908&0.851&7.035&22.505&0.743&12.166&23.820&0.759&10.691&42.8\cr
CAIN~\cite{choi2020channel}&DBPN~\cite{haris2019deep}&31.741&0.954&3.904&31.796&0.946&3.819&27.814&0.901&6.105&28.100&0.877&6.125&23.672&0.779&10.561&24.764&0.784&9.478&42.8+10.4\cr
CAIN~\cite{choi2020channel}&RBPN~\cite{haris2019recurrent}&31.721&0.955&3.896&31.980&0.949&3.702&27.995&0.906&5.930&28.377&0.887&5.855&23.566&0.781&10.498&24.605&0.787&9.437&42.8+12.7\cr
CAIN~\cite{choi2020channel}&EDVR~\cite{wang2019edvr}&---&---&---&---&---&---&28.339&0.911&5.711&28.690&0.893&5.642&---&---&---&---&---&---&42.8+20.7\cr
\hline

\multicolumn{2}{c|}{STARnet~\cite{haris2020space}}&---&---&---&---&---&---&28.829&0.920&---&30.608&0.926&---&---&---&---&---&---&---&111.6\cr
\multicolumn{2}{c|}{Zooming Slow-Mo~\cite{xiang2020zooming}}&32.200&\textcolor{blue}{0.959}&3.630&33.270&0.963&2.982&28.931&0.923&5.184&30.621&\textcolor{blue}{0.927}&4.354&25.376&0.850&8.054&26.829&0.851&7.018&11.1\cr
\multicolumn{2}{c|}{TMNet~\cite{xu2021temporal}}&\textcolor{blue}{32.211}&\textcolor{red}{0.960}&\textcolor{blue}{3.620}&\textcolor{blue}{33.298}&\textcolor{blue}{0.964}&\textcolor{blue}{2.974}&\textcolor{blue}{28.988}&\textcolor{blue}{0.924}&\textcolor{blue}{5.149}&\textcolor{blue}{30.699}&\textcolor{red}{0.929}&\textcolor{blue}{4.311}&\textcolor{blue}{25.424}&\textcolor{blue}{0.852}&\textcolor{blue}{7.984}&\textcolor{blue}{26.994}&\textcolor{blue}{0.854}&\textcolor{blue}{6.874}&12.3\cr
\multicolumn{2}{c|}{CycMu-Net}&\textcolor{red}{32.258}&\textcolor{red}{0.960}&\textcolor{red}{3.608}&\textcolor{red}{33.545}&\textcolor{red}{0.965}&\textcolor{red}{2.885}&\textcolor{red}{29.020}&\textcolor{red}{0.925}&\textcolor{red}{5.130}&\textcolor{red}{30.750}&\textcolor{red}{0.929}&\textcolor{red}{4.287}&\textcolor{red}{25.486}&\textcolor{red}{0.853}&\textcolor{red}{7.931}&\textcolor{red}{27.062}&\textcolor{red}{0.856}&\textcolor{red}{6.827}&11.1\cr
\hline
\end{tabular}
}
     \vspace{-2mm}
\caption{Quantitative comparisons ($\times$2, $\times$4, $\times$8 from left to right) of the state-of-the art methods for ST-VSR. The numbers in {\color{red}red} and {\color{blue}blue} represent the best and second best performance.}
   \vspace{-3mm}
\label{ALL-Eva}
\end{table*}

\tabcolsep=0.5pt
\begin{figure*}[t]
	\centering
\footnotesize{
		\begin{tabular}{ccccccc}
		    \includegraphics[width=0.2\textwidth]{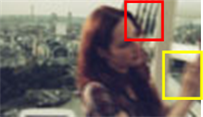} &
			\includegraphics[width=0.2\textwidth]{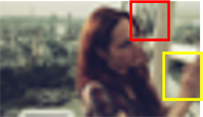} &
			\includegraphics[width=0.2\textwidth]{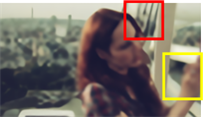} &
			\includegraphics[width=0.2\textwidth]{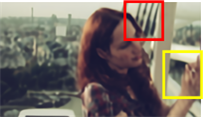} &
			\includegraphics[width=0.2\textwidth]{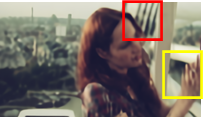} \\
			\includegraphics[width=0.2\textwidth]{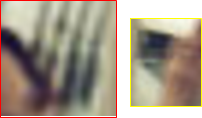} &
			\includegraphics[width=0.2\textwidth]{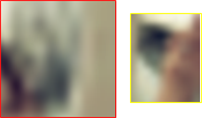} &
			\includegraphics[width=0.2\textwidth]{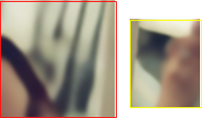} &
			\includegraphics[width=0.2\textwidth]{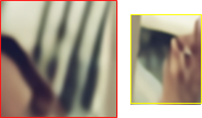} &
			\includegraphics[width=0.2\textwidth]{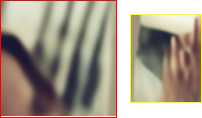} \\
			\textbf{Overlayed LR} & \textbf{AdaCoF+Bicubic} & \textbf{AdaCoF+DBPN} &\textbf{SepConv+RBPN} &\textbf{AdaCoF+RBPN}\\
			&\textbf{(24.425/0.804)}&\textbf{(26.995/0.858)}&\textbf{(28.383/0.885)}&\textbf{(27.835/0.877)}&\\
			\includegraphics[width=0.2\textwidth]{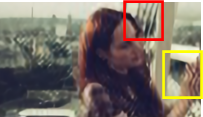} &
			\includegraphics[width=0.2\textwidth]{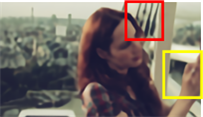} &
			\includegraphics[width=0.2\textwidth]{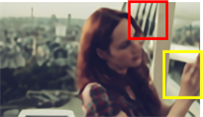} &
			\includegraphics[width=0.2\textwidth]{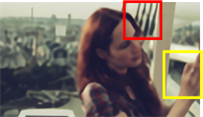} &
			\includegraphics[width=0.2\textwidth]{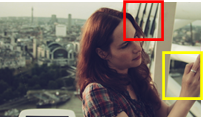} \\
			\includegraphics[width=0.2\textwidth]{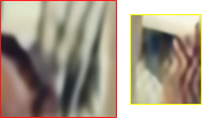} &
			\includegraphics[width=0.2\textwidth]{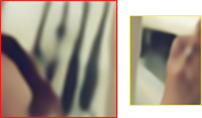} &
			\includegraphics[width=0.2\textwidth]{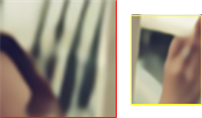} &
			\includegraphics[width=0.2\textwidth]{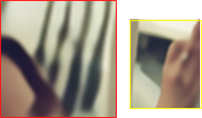} &
			\includegraphics[width=0.2\textwidth]{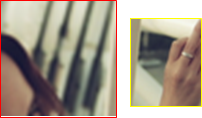} \\
			\textbf{CAIN+RBPN} & \textbf{Zooming Slow-Mo} & \textbf{TMNet} &\textbf{CycMu-Net} &\textbf{Ground-Truth}\\
			\textbf{(25.700/0.810)}&\textbf{(28.582/0.887)}&\textbf{\textcolor{blue}{(28.829/0.890)}}&\textbf{\textcolor{red}{(28.983/0.893)}}&\\
			
	\end{tabular}}
         \vspace{-3mm}
\caption{\small Visual comparisons (8$\times$) with state-of-the-art methods on \textbf{Vimeo90K} dataset.}
\vspace{-4mm}
\label{x8 visual}
\end{figure*}

\tabcolsep=0.5pt
\begin{figure*}[t]
	\centering
\footnotesize{
		\begin{tabular}{ccccccc}
		    \includegraphics[width=0.13\textwidth]{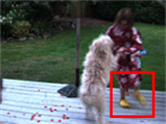} &
			\includegraphics[width=0.13\textwidth]{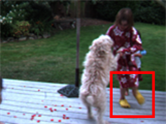} &
			\includegraphics[width=0.13\textwidth]{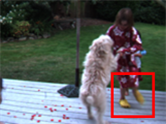} &
			\includegraphics[width=0.13\textwidth]{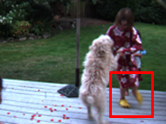} &
			\includegraphics[width=0.13\textwidth]{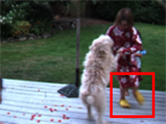} &
			\includegraphics[width=0.13\textwidth]{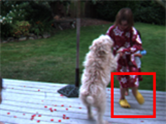}&
			\includegraphics[width=0.13\textwidth]{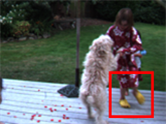}\\
			\includegraphics[width=0.13\textwidth]{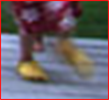} &
			\includegraphics[width=0.13\textwidth]{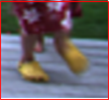} &
			\includegraphics[width=0.13\textwidth]{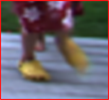} &
			\includegraphics[width=0.13\textwidth]{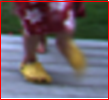} &
			\includegraphics[width=0.13\textwidth]{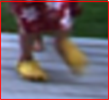}&
			\includegraphics[width=0.13\textwidth]{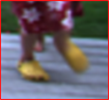}&
			\includegraphics[width=0.13\textwidth]{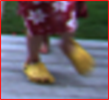}\\
			\textbf{Overlayed LR} & \textbf{EDSC} & \textbf{DAIN} &\textbf{AdaCoF++} &\textbf{CAIN}&\textbf{CycMu-Net} & \textbf{Ground-truth} \\
			&(1.826)&(1.868)&\textbf{\textcolor{blue}{(1.812)}}&(3.054)&\textbf{\textcolor{red}{(1.539)}}&\\
	\end{tabular}}
         \vspace{-4mm}
\caption{\small Visual comparisons of temporal video super-resolution on \textbf{Middlebury} dataset.}
   \vspace{-4mm}
\label{T-VSR visual}
\end{figure*}

\tabcolsep=0.5pt
\begin{figure*}[t]
	\centering
\footnotesize{
		\begin{tabular}{ccccccc}
		    \includegraphics[width=0.155\textwidth]{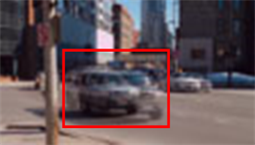} &
			\includegraphics[width=0.155\textwidth]{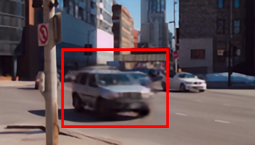} &
			\includegraphics[width=0.155\textwidth]{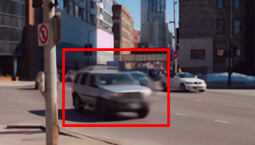} &
			\includegraphics[width=0.155\textwidth]{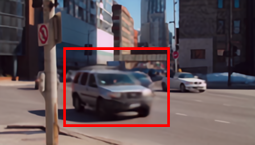} &
			\includegraphics[width=0.155\textwidth]{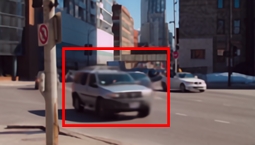} &
			\includegraphics[width=0.155\textwidth]{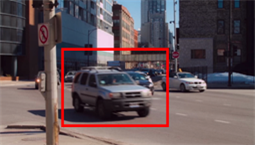}\\
			\includegraphics[width=0.155\textwidth]{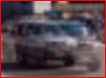} &
			\includegraphics[width=0.155\textwidth]{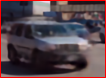} &
			\includegraphics[width=0.155\textwidth]{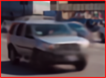} &
			\includegraphics[width=0.155\textwidth]{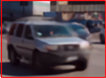} &
			\includegraphics[width=0.155\textwidth]{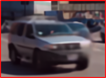} &
			\includegraphics[width=0.155\textwidth]{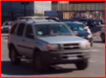} \\
			\textbf{Overlayed LR} & \textbf{Model (a)} & \textbf{Model (b)} &\textbf{Model (c)} &\textbf{Model (d)} & \textbf{Ground-truth} \\
			& \textbf{(27.182/0.887)} & \textbf{(27.276/0.893)} &\textbf{\textcolor{blue}{(27.528/0.897)}} &\textbf{\textcolor{red}{(27.673/0.900)}} & \\
			
	\end{tabular}}
         \vspace{-3mm}
\caption{\small Visual comparisons (4$\times$) of four variants for the ablation studies on \textbf{Vimeo90K} dataset.}
   \vspace{-3mm}
\label{variant visual}
\end{figure*}


\heading{UCF101~\cite{soomro2012ucf101}.}
The UCF101 dataset consists of videos with a large variety of human actions. There are 379 triplets with the resolution of 256$\times$256 for testing. \jk{The original images are sampled} to 32$\times$32, 64$\times$64 and 128$\times$128 with Bicubic 
for 8$\times$, 4$\times$ and 2$\times$ SR tasks in testing.

\heading{Middlebury~\cite{baker2011database}.}
The Middlebury dataset is widely used to evaluate video frame interpolation algorithms~\cite{bao2019depth,cheng2020video}. Here, we select Other set which provides the ground-truth middle frames, only to test our method on T-VSR task. The image resolution in this dataset is around 640$\times$480 pixels.

\heading{Metric.}
We use Peak Signal-to-Noise Ratio (PSNR), Structural Similarity Index (SSIM)~\cite{wang2004image} and the average Interpolation Error (IE) for performance evaluation. The higher PSNR and SSIM and lower IE \jk{values indicate} better super-resolution and interpolation performance.

\subsection{Comparisons with State-of-the-Art Methods}
\heading{ST-VSR.}
We compare our CycMu-Net with state-of-the-art two-stage and one-stage based ST-VSR methods. For the two-stage based ST-VSR methods, 
SepConv~\cite{niklaus2017videob}, AdaCoF~\cite{lee2020adacof} and CAIN~\cite{choi2020channel} \jk{are introduced for T-VSR task}, \jk{while} Bicubic Interpolation, RBPN~\cite{haris2019recurrent}, DBPN~\cite{haris2019deep} and EDVR~\cite{wang2019edvr} are used for S-VSR. For one-stage based ST-VSR methods, we compare our CycMu-Net with Zooming SlowMo~\cite{xiang2020zooming}, STARnet~\cite{haris2020space} and TMNet~\cite{xu2021temporal}. 
\jk{For fair comparison, 
three triplets from \textbf{Vimeo90K} dataset are used to retrain SlowMo and TMNet methods.} 

\heading{Quantitative results.} Quantitative results are presented in Table~\ref{ALL-Eva}. We can see that \jk{besides fewer parameters, one-stage based methods} 
\jk{show significant superiority} than the two-stage based methods in all metrics. \jk{In particular}, the best two-stage based method (SepConv+RBPN) is 0.66dB lower than our method for 8$\times$ VSR on Vimeo90K dataset. Furthermore, compared to the state-of-the-art one-stage based methods, 
our proposed CycMu-Net outperforms STARNet~\cite{haris2020space}, Zooming Slow-Mo~\cite{xiang2020zooming} and TMNet ~\cite{xu2021temporal} \jk{on all datasets with all metrics, while with only one-tenth of parameters to STARnet.} 
All these results validate the effectiveness of our proposed method for ST-VSR task. 

\heading{Qualitative results.} The qualitative results of seven ST-VSR baselines with their PSNR and SSIM values are shown in Figure~\ref{x8 visual}. 
\jk{As expected,} two-stage based ST-VSR methods tend to produce blurry results (see the yellow boxes) \jk{since} they ignore the mutual relations between S-VSR and T-VSR, \jk{which help the accurate texture inference}. Compared to two-stage based methods, one-stage based ST-VSR methods can generate complete results. However, these methods ignore that S-VSR provides abundant spatial information for the refinement of temporal prediction, leading to the generated results without more texture information (see red and yellow boxes). On the contrary, our proposed method explores the mutual relations between S-VSR and T-VSR, \jk{which contribute to generating} sharper results with clearer structure and texture. (More visual comparisons are provided in the supplementary materials)

\begin{table}[t]
\setlength\tabcolsep{8pt}

\centering
\resizebox{1.0\columnwidth}{!}
{
\smallskip\begin{tabular}{c|ccc|ccc|c}
\hline
\multirow{2}{*}{Methods}&\multicolumn{3}{c|}{UCF101} & \multicolumn{3}{c|}{Vimeo90K} &{Parameters}\cr
&PSNR&SSIM&IE&PSNR&SSIM&IE&(millions)\cr \hline
Bicubic&27.254&0.889&6.232&28.135&0.879&5.994&---\cr
DBPN~\cite{haris2019deep}&30.898&0.938&4.211&31.484&0.928&4.137&10.4\cr
RBPN~\cite{haris2019recurrent}&31.309&\textcolor{blue}{0.943}&4.035&32.417&0.939&3.759&12.7\cr
EDVR~\cite{wang2019edvr}&\textcolor{blue}{31.452}&\textcolor{red}{0.944}&\textcolor{red}{3.974}&\textcolor{red}{32.558}&\textcolor{red}{0.941}&\textcolor{red}{3.680}&20.7\cr
CycMu-Net~&\textcolor{red}{31.463}&\textcolor{red}{0.944}&\textcolor{blue}{3.980}&\textcolor{blue}{32.472}&\textcolor{blue}{0.940}&\textcolor{blue}{3.735}&11.1\cr

\hline
\end{tabular}
}
\vspace{-3mm} 
\caption{Quantitative comparisons of the state-of-the art methods for S-VSR ($H_{0}$) \jk{on UCF101 and Vimeo90K} datasets.}
\vspace{-6mm} 
\label{SVSR-Eva}
\end{table}



\heading{S-VSR.}
We compare the proposed network with image SR methods including Bicubic and DBPN~\cite{haris2019deep}, and S-VSR methods including RBPN~\cite{haris2019recurrent} and EDVR~\cite{wang2019edvr}. The results on S-VSR  are shown in Table~\ref{SVSR-Eva}, showing that S-VSR methods (EDVR~\cite{wang2019edvr} and RBPN~\cite{haris2019recurrent}) can achieve superior performance than image SR methods (bicubic and DBPN~\cite{haris2019deep})  by referring to multiple frames for temporal correlations. In addition, we can see that our CycMu-Net has comparable results with EDVR, but it requires only half of the parameters of EDVR and three triplets rather than seven frames for training. This also validates the powerful generalization ability of our network, and our proposed up-projection units are helpful for S-VSR tasks by exploiting temporal correlations from T-VSR.

\begin{table}[t]
\setlength\tabcolsep{3pt}
\centering
\resizebox{1.0\columnwidth}{!}
{
\smallskip\begin{tabular}{c|ccc|ccc|c|c}
\hline
\multirow{2}{*}{Methods}&\multicolumn{3}{c|}{UCF101} & \multicolumn{3}{c|}{Vimeo90K}&\multicolumn{1}{c|}{MB-Other} &{Parameters}\cr
&PSNR&SSIM&IE&PSNR&SSIM&IE&IE&(millions)\cr \hline
SpeConv-$L_{f}$~\cite{niklaus2017videob}&37.883&0.982&2.264&36.506&0.985&1.936&1.355&21.6\cr
SpeConv-$L_{1}$~\cite{niklaus2017videob}&37.953&\textcolor{blue}{0.983}&2.221&36.788&0.986&1.845&1.310&21.6\cr
EDSC~\cite{cheng2021multiple}&37.946&\textcolor{blue}{0.983}&2.271&\textcolor{blue}{37.326}&\textcolor{blue}{0.988}&\textcolor{blue}{1.824}&\textcolor{blue}{1.302}&8.9\cr
DAIN~\cite{bao2019depth}&38.172&\textcolor{blue}{0.983}&2.131&36.686&0.986&1.862&1.346&24.0\cr
CyclicGen++~\cite{liu2019deep}&37.644&0.981&2.261&33.935&0.973&2.660&1.750&19.8\cr
AdaCoF++~\cite{lee2020adacof}&\textcolor{blue}{38.387}&\textcolor{blue}{0.983}&\textcolor{blue}{2.088}&36.874&0.987&1.857&1.304&21.8\cr
CAIN~\cite{choi2020channel}&35.407&0.979&2.849&34.857&0.979&2.729&2.369&42.8\cr
CycMu-Net&\textcolor{red}{38.850}&\textcolor{red}{0.984}&\textcolor{red}{2.012}&\textcolor{red}{39.074}&\textcolor{red}{0.990}&\textcolor{red}{1.422}&\textcolor{red}{0.983}&11.1\cr
\hline
\end{tabular}
}
   \vspace{-2mm}
\caption{Quantitative comparisons of the state-of-the art methods for T-VSR ($L_{t}$).}
   \vspace{-6mm}
\label{TVSR-Eva}
\end{table}

\heading{T-VSR.}
We compare our proposed network with state-of-the-art T-VSR which include SpeConv-$L_{f}$~\cite{niklaus2017videob}, SepConv-$L_{1}$~\cite{niklaus2017videob},  EDSC~\cite{cheng2021multiple}, DAIN~\cite{bao2019depth}, CyclicGen++~\cite{liu2019deep}, AdaCoF++~\cite{lee2020adacof} and CAIN~\cite{choi2020channel}. The results on T-VSR are shown in Table~\ref{TVSR-Eva}. We can find that our proposed method is significantly better than the state-of-the-art video frame interpolation. For example, PSNR values of our proposed CycMu-Net 
are 1.1dB and 1.6dB higher than EDSC~\cite{cheng2021multiple} on UCF101 and Vimeo90K datasets, respectively. In addition, we show the visualized results and IE value from four temporal video super-resolution method in Figure~\ref{T-VSR visual}, our proposed method produces intermediate frame with more details (\textit{e.g.}, the shoe). We attribute this to the fact that when we train the ST-VSR network, we make full use of HR information from S-VSR via down-projection units. Therefore, the interpolated frame can obtain more texture and detailed information from S-VSR.

\tabcolsep=0.5pt
\begin{figure*}[htb]
	\centering
\footnotesize{
		\begin{tabular}{ccccccc}
		    \includegraphics[width=0.135\textwidth]{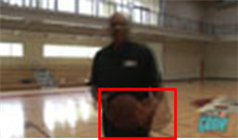} &
			\includegraphics[width=0.135\textwidth]{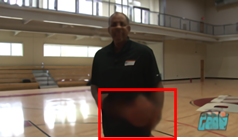} &
			\includegraphics[width=0.135\textwidth]{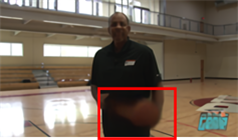} &
			\includegraphics[width=0.135\textwidth]{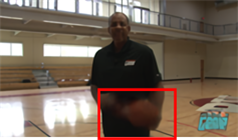} &
			\includegraphics[width=0.135\textwidth]{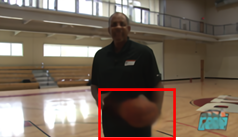} &
			\includegraphics[width=0.135\textwidth]{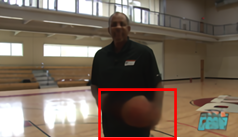} &
			\includegraphics[width=0.135\textwidth]{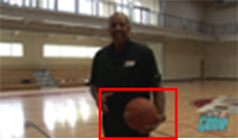}\\
			\includegraphics[width=0.135\textwidth]{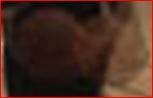} &
			\includegraphics[width=0.135\textwidth]{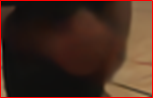} &
			\includegraphics[width=0.135\textwidth]{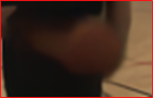} &
			\includegraphics[width=0.135\textwidth]{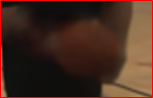} &
			\includegraphics[width=0.135\textwidth]{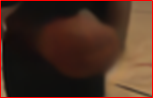} &
			\includegraphics[width=0.135\textwidth]{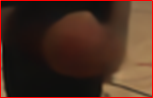}&
			\includegraphics[width=0.135\textwidth]{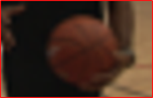}\\
			\textbf{Overlayed LR} & \textbf{M=2} & \textbf{M=4} &\textbf{M=6} &\textbf{M=8} & \textbf{M=10} & \textbf{Ground-truth} \\
			& \textbf{(26.095/0.897)} & \textbf{(26.195/0.901)} &\textbf{(26.214/0.901)} &\textbf{\textcolor{blue}{(26.229/0.902)}} & \textbf{\textcolor{red}{(26.330/0.904)}} & \\
	\end{tabular}}
         \vspace{-3mm}
\caption{\small Visual comparisons (4$\times$) of different numbers of up-projection and down-projection units for the ablation studies on \textbf{Vimeo90K} dataset.}
   \vspace{-4mm}
\label{units visual}
\end{figure*}

\subsection{Model Analysis}
\heading{Ablation Study.}
To further verify 
the key modules in 
CycMu-Net, comprehensive ablation studies are conducted for 4$\times$ SR.

\begin{table}[t]
\smallskip
\centering
\setlength\tabcolsep{2pt}
\resizebox{\columnwidth}{!}
{
\begin{tabular}{c|cccc|ccc|ccc}
\hline
\multirow{2}{*}{Methods}&\multicolumn{2}{c}{FTI}&\multicolumn{2}{c|}{PU}&
\multicolumn{3}{c|}{UCF101}&\multicolumn{3}{c}{Vimeo90K}\cr
~&FFI&DFI&PP&CP&PSNR&SSIM&IE&PSNR&SSIM&IE\\
\hline
\textbf{Model (a)}&\cmark&&&&28.861&\textcolor{blue}{0.922}&5.243&30.170&0.921&4.616\cr
\textbf{Model (b)}&&\cmark&&&28.926&\textcolor{red}{0.924}&5.161&30.510&\textcolor{blue}{0.926}&4.415\cr
\hline
\textbf{Model (c)}&&\cmark&\cmark&&\textcolor{blue}{28.940}&\textcolor{red}{0.924}&\textcolor{blue}{5.150}&\textcolor{blue}{30.544}&\textcolor{blue}{0.926}&\textcolor{blue}{4.390}\cr
\textbf{Model (d)}&&\cmark&&\cmark&\textcolor{red}{28.996}&\textcolor{red}{0.924}&\textcolor{red}{5.144}&\textcolor{red}{30.650}&\textcolor{red}{0.928}&\textcolor{red}{4.338}\cr
\hline
\end{tabular}
}
\vspace{-2mm} 
\caption{Quantitative comparisons on the performance (4$\times$) of different modules. FTI denotes feature temporal interpolation, FFI denotes fusion feature interpolation, DFI denotes deformable feature interpolation, PU denotes projection units. PP denotes plain-projected units and CP denotes cycle-projected units.}
\vspace{-4mm}
\label{variant ablation}
\end{table}

\heading{Model (a)}: \jk{A fusion feature interpolation (FFT) network is used to direct fuse input information from input frames and produce intermediate representation without motion estimation.} 
Then two pixel-shuffle layers take the representations as inputs, and produce the 4$\times$ SR video with a convolution.

\heading{Model (b)}: \Hu{We add deformable convolution as implicit motion estimation into feature interpolation network (FTI-Net) in Model (a) as our deformable feature interpolation (DFI) network, as stated in section 3.2} 

\heading{Model (c)}: Based on Model (b), we add addition iterative plain-projection units (PP) \Hu{without up-down sampling} between the feature temporal interpolation network and reconstruction network.

\heading{Model (d)}: 
\jk{The complete version} of CycMu-Net.

The visual and numerical comparisons are shown in Figure~\ref{variant visual} and Table~\ref{variant ablation}. Compared to Model (a) that produces the intermediate representations without motion estimation, the results of Model (b) show that adopting deformable convolution for implicit frame interpolation can bring 0.34dB gain on Vimeo90K dataset and improves the visual result (\textit{e.g.}, the edge of the moving car). Based on Model (b), the addition of plain projection units (Model (c)) can help Model (b) to generate a car with clearer structure. Unfortunately, they fail to recover key details (\textit{e.g.}, license plate). On the contrary, our proposed Model (d) can generate more credible SR results. It demonstrates the fact that our proposed up- and down-projection units eliminate cross-space errors while plain-projection units magnify errors.

\begin{table}[t]
\setlength\tabcolsep{6pt}
\centering
\resizebox{1.0\columnwidth}{!}
{
\smallskip\begin{tabular}{c|ccc|ccc|c}
\hline
\multirow{2}{*}{M}&\multicolumn{3}{c|}{UCF101} & \multicolumn{3}{c|}{Vimeo90K} &{Parameters}\cr
&PSNR&SSIM&IE&PSNR&SSIM&IE&(millions)\cr \hline
2&28.939&{0.923}&5.181&30.480&0.926&4.420&7.3\cr
4&28.982&\textcolor{blue}{0.924}&5.149&30.601&\textcolor{blue}{0.927}&4.360&9.2\cr
6&29.020&\textcolor{red}{0.925}&\textcolor{blue}{5.130}&30.750&\textcolor{red}{0.929}&4.287&11.1\cr
8&\textcolor{blue}{29.030}&\textcolor{red}{0.925}&\textcolor{blue}{5.130}&\textcolor{blue}{30.753}&\textcolor{red}{0.929}&\textcolor{blue}{4.282}&13.0\cr
10&\textcolor{red}{29.044}&\textcolor{red}{0.925}&\textcolor{red}{5.128}&\textcolor{red}{30.791}&\textcolor{red}{0.929}&\textcolor{red}{4.273}&14.9\cr
\hline
\end{tabular}
}
   \vspace{-2mm}
\caption{Quantitative comparisons on the performance (4$\times$) of different number of projection units.}
   \vspace{-4mm}
\label{M-ablation}
\end{table}

\begin{figure}[t]
	\centering
\footnotesize{
		\begin{tabular}{ccccccc}
		    \includegraphics[width=0.245\linewidth]{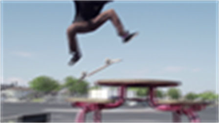} &
			\includegraphics[width=0.245\linewidth]{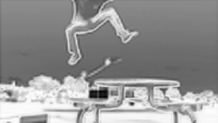} &
			\includegraphics[width=0.245\linewidth]{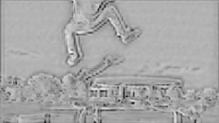} &
			\includegraphics[width=0.245\linewidth]{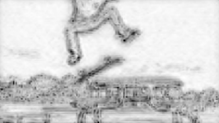} \\
			\textbf{Overlayed LR} & \textbf{$h_{t}^{1}$} & \textbf{$h_{t}^{2}$ } &\textbf{ $h_{t}^{3}$}\\
			\includegraphics[width=0.245\linewidth]{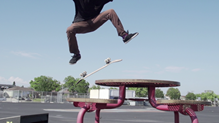} &
			\includegraphics[width=0.245\linewidth]{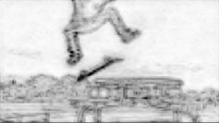} &
			\includegraphics[width=0.245\linewidth]{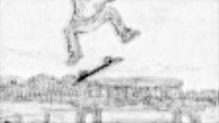} &
			\includegraphics[width=0.245\linewidth]{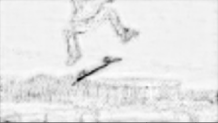}\\
			\textbf{Ground-truth}&\textbf{$h_{t}^{4}$} & \textbf{$h_{t}^{5}$} & \textbf{ $h_{t}^{6}$} \\
	\end{tabular}}
         \vspace{-3mm}
\caption{\small Feature maps from up-projection units in CycMu-Net where $M=6$. Each feature map has been visualized using same grayscale colormap.}
   \vspace{-6mm}
\label{feature visual}
\end{figure}

\heading{Impacts of Up-projection and Down-projection Units.}
To demonstrate the effectiveness of our up-projection units and down-projection units, we construct multiple networks ($M=2,4,6,8,10$) by setting different numbers of projection units. The visual and numerical results on 4 $\times$ are shown in Figure~\ref{units visual} and Table~\ref{M-ablation}. As the numbers of up-projection and down-projection units increase, CycMu-Net produces results with more complete structure and details (\textit{e.g.}, the basketball), and achieves better results in term of PSNR, SSIM and IE on two datasets. Considering the trade-off between efficacy and efficiency, we set $M$ to 6 to predict the final results of the proposed CycMu-Net. These also verify that the proposed up-projection and down-projection units play important roles in mutually benefiting from S-VSR and T-VSR. In addition, in order to analyze the specific role of the projection units that temporal correlations are exploited to promote the texture and detail information. In Figure~\ref{feature visual}, it is shown that each up-projection unit generates feature map, which contains different types of HR components and increases the quality of S-VSR. This demonstrates that multiple up-projection units can obtain diverse HR representations for guiding the better super-resolution reconstruction.

\tabcolsep=0.5pt

\section{Conclusion}
In this work, we propose a novel one-stage based Cycle-projected Mutual learning network (CycMu-Net) for spatial-temporal video super-resolution. 
\jk{Theoretically, we introduce mutual learning 
to explore the interactions between spatial video super-resolution (S-VSR) and temporal video super-resolution (T-VSR), from which the abundant spatial information and temporal correlations are aggregated to infer accurate intermediate frame. Specifically, an elaborate iterative representation between up-projection units and down-projection units is introduced to make full use of the spatial-temporal features while eliminating the inference errors.} 
Extensive experiments 
demonstrate our proposed method performs well against the state-of-the-art methods in \jk{both S-VSR, T-VSR and ST-VSR} tasks. 
\jk{While achieving impressive performance, one limitation of this study is that since videos might contain dramatically changing scenes, the spatial-temporal correlations of large motion or SR factors is hardly predicted via the iterative up-projection and down-projection units. One reasonable scheme is to alleviate the learning burden by dividing it into multiple sub-tasks with small motion, which is helpful for accurate texture inference.}

\heading{Acknowledgements}.
This work was supported by National Key R\&D Project (2021YFC3320301) and National Natural Science Foundation of China (62171325). The numerical calculations in this paper have been done on the supercomputing system in the Supercomputing Center of Wuhan University.

{\small
\bibliographystyle{ieee_fullname}
\bibliography{egbib}

\begin{thebibliography}{10}\itemsep=-1pt

\bibitem{baker2011database}
Simon Baker, Daniel Scharstein, JP Lewis, Stefan Roth, Michael~J Black, and
  Richard Szeliski.
\newblock A database and evaluation methodology for optical flow.
\newblock {\em IJCV}, 92(1):1--31, 2011.

\bibitem{bao2019depth}
Wenbo Bao, Wei-Sheng Lai, Chao Ma, Xiaoyun Zhang, Zhiyong Gao, and Ming-Hsuan
  Yang.
\newblock Depth-aware video frame interpolation.
\newblock In {\em CVPR}, pages 3703--3712, 2019.

\bibitem{bao2019memc}
Wenbo Bao, Wei-Sheng Lai, Xiaoyun Zhang, Zhiyong Gao, and Ming-Hsuan Yang.
\newblock Memc-net: Motion estimation and motion compensation driven neural
  network for video interpolation and enhancement.
\newblock {\em IEEE TPAMI}, 2019.

\bibitem{batra2017cooperative}
Tanmay Batra and Devi Parikh.
\newblock Cooperative learning with visual attributes.
\newblock {\em arXiv preprint arXiv:1705.05512}, 2017.

\bibitem{boykov2001fast}
Yuri Boykov, Olga Veksler, and Ramin Zabih.
\newblock Fast approximate energy minimization via graph cuts.
\newblock {\em IEEE TPAMI}, 23(11):1222--1239, 2001.

\bibitem{caballero2017real}
Jose Caballero, Christian Ledig, Andrew Aitken, Alejandro Acosta, Johannes
  Totz, Zehan Wang, and Wenzhe Shi.
\newblock Real-time video super-resolution with spatio-temporal networks and
  motion compensation.
\newblock In {\em CVPR}, pages 4778--4787, 2017.

\bibitem{chan2020understanding}
Kelvin~CK Chan, Xintao Wang, Ke Yu, Chao Dong, and Chen~Change Loy.
\newblock Understanding deformable alignment in video super-resolution.
\newblock {\em arXiv preprint arXiv:2009.07265}, 4:3, 2020.

\bibitem{chan2021basicvsr}
Kelvin~CK Chan, Xintao Wang, Ke Yu, Chao Dong, and Chen~Change Loy.
\newblock Basicvsr: The search for essential components in video
  super-resolution and beyond.
\newblock In {\em CVPR}, pages 4947--4956, 2021.

\bibitem{charbonnier1994two}
Pierre Charbonnier, Laure Blanc-Feraud, Gilles Aubert, and Michel Barlaud.
\newblock Two deterministic half-quadratic regularization algorithms for
  computed imaging.
\newblock In {\em ICIP}, volume~2, pages 168--172. IEEE, 1994.

\bibitem{cheng2020video}
Xianhang Cheng and Zhenzhong Chen.
\newblock Video frame interpolation via deformable separable convolution.
\newblock In {\em AAAI}, volume~34, pages 10607--10614, 2020.

\bibitem{cheng2021multiple}
Xianhang Cheng and Zhenzhong Chen.
\newblock Multiple video frame interpolation via enhanced deformable separable
  convolution.
\newblock {\em IEEE TPAMI}, 2021.

\bibitem{choi2020channel}
Myungsub Choi, Heewon Kim, Bohyung Han, Ning Xu, and Kyoung~Mu Lee.
\newblock Channel attention is all you need for video frame interpolation.
\newblock In {\em AAAI}, volume~34, pages 10663--10671, 2020.

\bibitem{dai2017deformable}
Jifeng Dai, Haozhi Qi, Yuwen Xiong, Yi Li, Guodong Zhang, Han Hu, and Yichen
  Wei.
\newblock Deformable convolutional networks.
\newblock In {\em ICCV}, pages 764--773, 2017.

\bibitem{dai2007bilateral}
Shengyang Dai, Mei Han, Ying Wu, and Yihong Gong.
\newblock Bilateral back-projection for single image super resolution.
\newblock In {\em ICME}, pages 1039--1042. IEEE, 2007.

\bibitem{dong2020multi}
Hang Dong, Jinshan Pan, Lei Xiang, Zhe Hu, Xinyi Zhang, Fei Wang, and
  Ming-Hsuan Yang.
\newblock Multi-scale boosted dehazing network with dense feature fusion.
\newblock In {\em CVPR}, pages 2157--2167, 2020.

\bibitem{dong2009nonlocal}
Weisheng Dong, Lei Zhang, Guangming Shi, and Xiaolin Wu.
\newblock Nonlocal back-projection for adaptive image enlargement.
\newblock In {\em ICIP}, pages 349--352. IEEE, 2009.

\bibitem{geman1984stochastic}
Stuart Geman and Donald Geman.
\newblock Stochastic relaxation, gibbs distributions, and the bayesian
  restoration of images.
\newblock {\em IEEE TPAMI}, (6):721--741, 1984.

\bibitem{gui2020featureflow}
Shurui Gui, Chaoyue Wang, Qihua Chen, and Dacheng Tao.
\newblock Featureflow: Robust video interpolation via structure-to-texture
  generation.
\newblock In {\em CVPR}, pages 14004--14013, 2020.

\bibitem{haris2019deep}
Muhammad Haris, Greg Shakhnarovich, and Norimichi Ukita.
\newblock Deep back-projection networks for single image super-resolution.
\newblock {\em CVPR}, 2019.

\bibitem{haris2019recurrent}
Muhammad Haris, Gregory Shakhnarovich, and Norimichi Ukita.
\newblock Recurrent back-projection network for video super-resolution.
\newblock In {\em CVPR}, pages 3897--3906, 2019.

\bibitem{haris2020deep}
Muhammad Haris, Greg Shakhnarovich, and Norimichi Ukita.
\newblock Deep back-projectinetworks for single image super-resolution.
\newblock {\em IEEE TPAMI}, 43(12):4323--4337, 2020.

\bibitem{haris2020space}
Muhammad Haris, Greg Shakhnarovich, and Norimichi Ukita.
\newblock Space-time-aware multi-resolution video enhancement.
\newblock In {\em CVPR}, pages 2859--2868, 2020.

\bibitem{he2016dual}
Di He, Yingce Xia, Tao Qin, Liwei Wang, Nenghai Yu, Tie-Yan Liu, and Wei-Ying
  Ma.
\newblock Dual learning for machine translation.
\newblock {\em NIPS}, 29, 2016.

\bibitem{hu2020motion}
Mengshun Hu, Liang Liao, Jing Xiao, Lin Gu, and Shin’ichi Satoh.
\newblock Motion feedback design for video frame interpolation.
\newblock In {\em ICASSP}, pages 4347--4351. IEEE, 2020.

\bibitem{hu2021capturing}
Mengshun Hu, Jing Xiao, Liang Liao, Zheng Wang, Chia-Wen Lin, Mi Wang, and
  Shin’ichi Satoh.
\newblock Capturing small, fast-moving objects: Frame interpolation via
  recurrent motion enhancement.
\newblock {\em IEEE TCSVT}, 2021.

\bibitem{huang2017video}
Yan Huang, Wei Wang, and Liang Wang.
\newblock Video super-resolution via bidirectional recurrent convolutional
  networks.
\newblock {\em IEEE TPAMI}, 40(4):1015--1028, 2017.

\bibitem{hui2018liteflownet}
Tak-Wai Hui, Xiaoou Tang, and Chen~Change Loy.
\newblock Liteflownet: A lightweight convolutional neural network for optical
  flow estimation.
\newblock In {\em CVPR}, pages 8981--8989, 2018.

\bibitem{jiang2018super}
Huaizu Jiang, Deqing Sun, Varun Jampani, Ming-Hsuan Yang, Erik Learned-Miller,
  and Jan Kautz.
\newblock Super slomo: High quality estimation of multiple intermediate frames
  for video interpolation.
\newblock In {\em CVPR}, pages 9000--9008, 2018.

\bibitem{kang2020deep}
Jaeyeon Kang, Younghyun Jo, Seoung~Wug Oh, Peter Vajda, and Seon~Joo Kim.
\newblock Deep space-time video upsampling networks.
\newblock In {\em ECCV}, pages 701--717. Springer, 2020.

\bibitem{kim2020fisr}
Soo~Ye Kim, Jihyong Oh, and Munchurl Kim.
\newblock Fisr: deep joint frame interpolation and super-resolution with a
  multi-scale temporal loss.
\newblock In {\em AAAI}, volume~34, pages 11278--11286, 2020.

\bibitem{kingma2014adam}
Diederik~P Kingma and Jimmy Ba.
\newblock Adam: A method for stochastic optimization.
\newblock {\em arXiv preprint arXiv:1412.6980}, 2014.

\bibitem{lai2017deep}
Wei-Sheng Lai, Jia-Bin Huang, Narendra Ahuja, and Ming-Hsuan Yang.
\newblock Deep laplacian pyramid networks for fast and accurate
  super-resolution.
\newblock In {\em CVPR}, pages 624--632, 2017.

\bibitem{lee2020adacof}
Hyeongmin Lee, Taeoh Kim, Tae-young Chung, Daehyun Pak, Yuseok Ban, and
  Sangyoun Lee.
\newblock Adacof: Adaptive collaboration of flows for video frame
  interpolation.
\newblock In {\em CVPR}, pages 5316--5325, 2020.

\bibitem{li2020mucan}
Wenbo Li, Xin Tao, Taian Guo, Lu Qi, Jiangbo Lu, and Jiaya Jia.
\newblock Mucan: Multi-correspondence aggregation network for video
  super-resolution.
\newblock In {\em ECCV}, pages 335--351. Springer, 2020.

\bibitem{liu2019deep}
Yu-Lun Liu, Yi-Tung Liao, Yen-Yu Lin, and Yung-Yu Chuang.
\newblock Deep video frame interpolation using cyclic frame generation.
\newblock In {\em AAAI}, volume~33, pages 8794--8802, 2019.

\bibitem{mudenagudi2010space}
Uma Mudenagudi, Subhashis Banerjee, and Prem~Kumar Kalra.
\newblock Space-time super-resolution using graph-cut optimization.
\newblock {\em IEEE TPAMI}, 33(5):995--1008, 2010.

\bibitem{niklaus2018context}
Simon Niklaus and Feng Liu.
\newblock Context-aware synthesis for video frame interpolation.
\newblock In {\em CVPR}, pages 1701--1710, 2018.

\bibitem{niklaus2020softmax}
Simon Niklaus and Feng Liu.
\newblock Softmax splatting for video frame interpolation.
\newblock In {\em CVPR}, pages 5437--5446, 2020.

\bibitem{niklaus2017videoa}
Simon Niklaus, Long Mai, and Feng Liu.
\newblock Video frame interpolation via adaptive convolution.
\newblock In {\em CVPR}, pages 670--679, 2017.

\bibitem{niklaus2017videob}
Simon Niklaus, Long Mai, and Feng Liu.
\newblock Video frame interpolation via adaptive separable convolution.
\newblock In {\em ICCV}, pages 261--270, 2017.

\bibitem{ranjan2017optical}
Anurag Ranjan and Michael~J Black.
\newblock Optical flow estimation using a spatial pyramid network.
\newblock In {\em CVPR}, pages 4161--4170, 2017.

\bibitem{sajjadi2018frame}
Mehdi~SM Sajjadi, Raviteja Vemulapalli, and Matthew Brown.
\newblock Frame-recurrent video super-resolution.
\newblock In {\em CVPR}, pages 6626--6634, 2018.

\bibitem{shechtman2002increasing}
Eli Shechtman, Yaron Caspi, and Michal Irani.
\newblock Increasing space-time resolution in video.
\newblock In {\em ECCV}, pages 753--768. Springer, 2002.

\bibitem{shi2021video}
Zhihao Shi, Xiaohong Liu, Kangdi Shi, Linhui Dai, and Jun Chen.
\newblock Video frame interpolation via generalized deformable convolution.
\newblock {\em IEEE TMM}, 2021.

\bibitem{soomro2012ucf101}
Khurram Soomro, Amir~Roshan Zamir, and Mubarak Shah.
\newblock Ucf101: A dataset of 101 human actions classes from videos in the
  wild.
\newblock {\em arXiv preprint arXiv:1212.0402}, 2012.

\bibitem{sun2018pwc}
Deqing Sun, Xiaodong Yang, Ming-Yu Liu, and Jan Kautz.
\newblock Pwc-net: Cnns for optical flow using pyramid, warping, and cost
  volume.
\newblock In {\em CVPR}, pages 8934--8943, 2018.

\bibitem{tao2017detail}
Xin Tao, Hongyun Gao, Renjie Liao, Jue Wang, and Jiaya Jia.
\newblock Detail-revealing deep video super-resolution.
\newblock In {\em CVPR}, pages 4472--4480, 2017.

\bibitem{teed2020raft}
Zachary Teed and Jia Deng.
\newblock Raft: Recurrent all-pairs field transforms for optical flow.
\newblock In {\em ECCV}, pages 402--419. Springer, 2020.

\bibitem{tian2020tdan}
Yapeng Tian, Yulun Zhang, Yun Fu, and Chenliang Xu.
\newblock Tdan: Temporally-deformable alignment network for video
  super-resolution.
\newblock In {\em CVPR}, pages 3360--3369, 2020.

\bibitem{wang2018learning}
Longguang Wang, Yulan Guo, Zaiping Lin, Xinpu Deng, and Wei An.
\newblock Learning for video super-resolution through hr optical flow
  estimation.
\newblock In {\em ACCV}, pages 514--529. Springer, 2018.

\bibitem{wang2019edvr}
Xintao Wang, Kelvin~CK Chan, Ke Yu, Chao Dong, and Chen Change~Loy.
\newblock Edvr: Video restoration with enhanced deformable convolutional
  networks.
\newblock In {\em CVPRW}, pages 0--0, 2019.

\bibitem{wang2004image}
Zhou Wang, Alan~C Bovik, Hamid~R Sheikh, and Eero~P Simoncelli.
\newblock Image quality assessment: from error visibility to structural
  similarity.
\newblock {\em TIP}, 13(4):600--612, 2004.

\bibitem{xiang2020zooming}
Xiaoyu Xiang, Yapeng Tian, Yulun Zhang, Yun Fu, Jan~P Allebach, and Chenliang
  Xu.
\newblock Zooming slow-mo: Fast and accurate one-stage space-time video
  super-resolution.
\newblock In {\em CVPR}, pages 3370--3379, 2020.

\bibitem{xingjian2015convolutional}
SHI Xingjian, Zhourong Chen, Hao Wang, Dit-Yan Yeung, Wai-Kin Wong, and
  Wang-chun Woo.
\newblock Convolutional lstm network: A machine learning approach for
  precipitation nowcasting.
\newblock In {\em NIPS}, pages 802--810, 2015.

\bibitem{xu2021temporal}
Gang Xu, Jun Xu, Zhen Li, Liang Wang, Xing Sun, and Ming-Ming Cheng.
\newblock Temporal modulation network for controllable space-time video
  super-resolution.
\newblock In {\em CVPR}, pages 6388--6397, 2021.

\bibitem{xue2019video}
Tianfan Xue, Baian Chen, Jiajun Wu, Donglai Wei, and William~T Freeman.
\newblock Video enhancement with task-oriented flow.
\newblock {\em IJCV}, 127(8):1106--1125, 2019.

\bibitem{yi2019progressive}
Peng Yi, Zhongyuan Wang, Kui Jiang, Junjun Jiang, and Jiayi Ma.
\newblock Progressive fusion video super-resolution network via exploiting
  non-local spatio-temporal correlations.
\newblock In {\em ICCV}, pages 3106--3115, 2019.

\bibitem{zhang2018image}
Yulun Zhang, Kunpeng Li, Kai Li, Lichen Wang, Bineng Zhong, and Yun Fu.
\newblock Image super-resolution using very deep residual channel attention
  networks.
\newblock In {\em ECCV}, pages 286--301, 2018.

\bibitem{zhang2018deep}
Ying Zhang, Tao Xiang, Timothy~M Hospedales, and Huchuan Lu.
\newblock Deep mutual learning.
\newblock In {\em CVPR}, pages 4320--4328, 2018.

\bibitem{zhao2017iterative}
Yang Zhao, Rong-Gang Wang, Wei Jia, Wen-Min Wang, and Wen Gao.
\newblock Iterative projection reconstruction for fast and efficient image
  upsampling.
\newblock {\em Neurocomputing}, 226:200--211, 2017.

\bibitem{zhu2019deformable}
Xizhou Zhu, Han Hu, Stephen Lin, and Jifeng Dai.
\newblock Deformable convnets v2: More deformable, better results.
\newblock In {\em CVPR}, pages 9308--9316, 2019.

\end{thebibliography}
}

\end{document}